\documentclass[11pt]{article}

\usepackage[final]{acl}

\usepackage{times}
\usepackage{latexsym}

\usepackage[T1]{fontenc}

\usepackage[utf8]{inputenc}

\usepackage{microtype}

\usepackage{inconsolata}

\usepackage{graphicx}
\usepackage{booktabs}
\usepackage{multicol}
\usepackage{multirow}
\usepackage{makecell}
\usepackage{url}
\usepackage{subcaption}
\usepackage{arydshln}

\usepackage[whole]{bxcjkjatype}

\title{CxMP: A Linguistic Minimal-Pair Benchmark for Evaluating Constructional Understanding in Language Models}

\author{%
Miyu Oba$^{1}$
~Saku Sugawara$^{2,3}$ \\
$^1$Nara Institute of Science and Technology \\
$^2$National Institute of Informatics ~$^3$The University of Tokyo \\
\texttt{oba.miyu.ol2@is.naist.jp} ~\texttt{saku@nii.ac.jp}
}

\begin{document}
\maketitle
\begin{abstract}
Recent work has examined language models from a linguistic perspective to better understand how they acquire language.
Most existing benchmarks focus on judging grammatical acceptability, whereas the ability to interpret meanings conveyed by grammatical forms has received much less attention.
We introduce the Linguistic Minimal-Pair Benchmark for Evaluating Constructional Understanding in Language Models (CxMP), a benchmark grounded in Construction Grammar that treats form–meaning pairings, or constructions, as fundamental linguistic units.
CxMP evaluates whether models can interpret the semantic relations implied by constructions, using a controlled minimal-pair design across nine construction types, including the let-alone, caused-motion, and ditransitive constructions.
Our results show that while syntactic competence emerges early, constructional understanding develops more gradually and remains limited even in large language models (LLMs). 
CxMP thus reveals persistent gaps in how language models integrate form and meaning, providing a framework for studying constructional understanding and learning trajectories in language models.

\end{abstract}

\section{Introduction}
Recent work increasingly analyzes language models from a linguistic perspective to better understand the sources of their success and their potential for language acquisition~\cite{hu-etal-2024-findings,kallini-etal-2024-mission,mita-etal-2025-developmentally}.
They primarily evaluate models in terms of their grammatical acceptability, that is, whether they distinguish grammatically acceptable from unacceptable sentences based on form-level constraints~\cite{warstadt-etal-2020-blimp-benchmark,warstadt-etal-2019-neural,weissweiler-etal-2025-linguistic}.
Such methods reflect a linguistic tradition that often treats syntax as largely independent from meaning, thereby focusing on the grammaticality of linguistic expressions.
However, language understanding also requires interpreting the meanings implied by form.
For instance, ``Mary ran into John.'' not only satisfies grammatical constraints but also conveys that ``Mary'' moved toward ``John'', who serves as the endpoint of that motion.
To bridge this gap, we adopt the framework of Construction Grammar, which treats form–meaning pairings (constructions) as the basic units of linguistic analysis.
In this view, syntax and meaning are inherently interconnected, enabling analyses that focus on meanings conveyed by linguistic form.

\begin{figure}[tbp]
\centering
\includegraphics[width=0.9\linewidth]{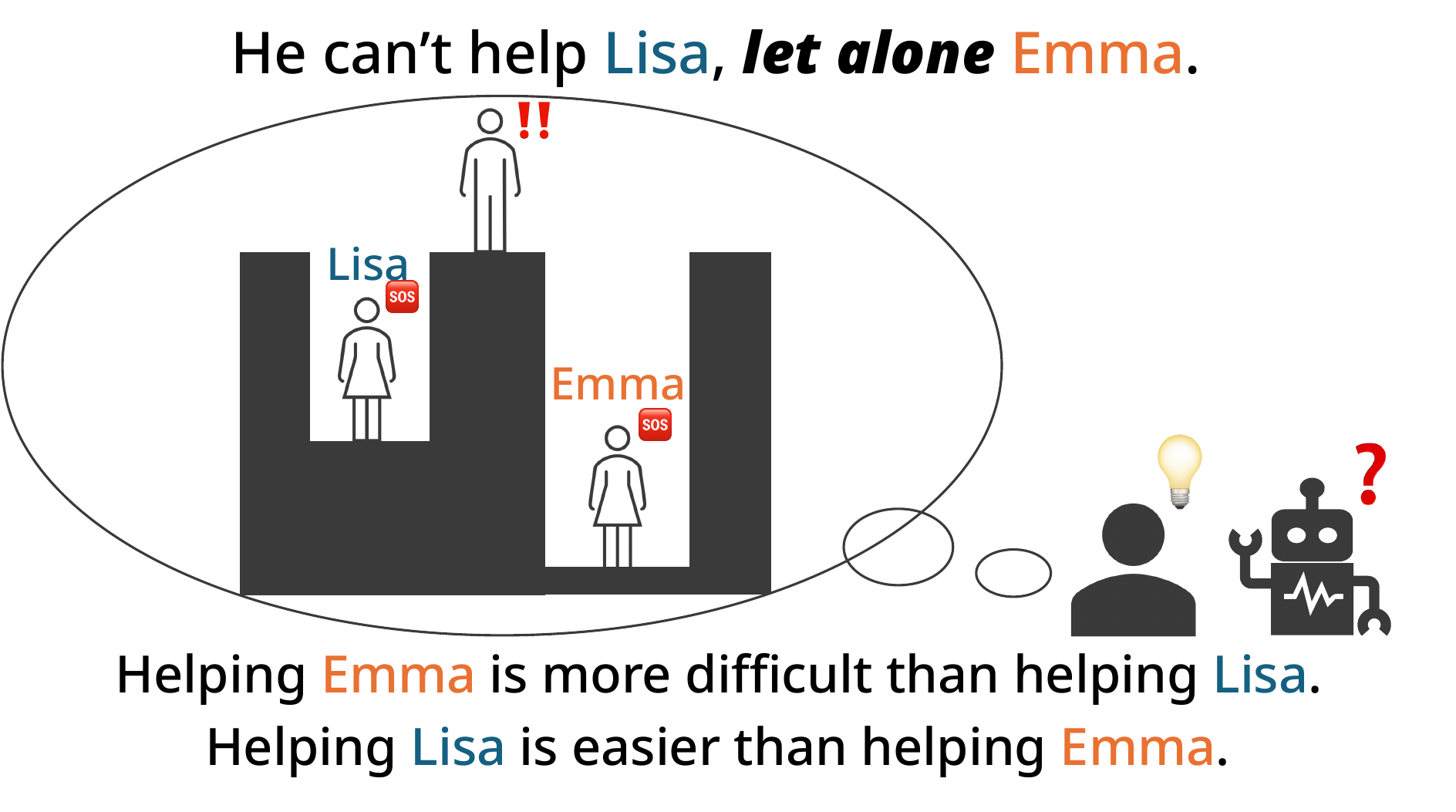}
\caption{Overview of the let-alone construction. Humans can grasp the relation between the two entities from sentences containing this construction.}
\label{fig:overview}
\end{figure}

Previous studies have evaluated language models from a constructional perspective~\cite{scivetti-etal-2025-beyond,weissweiler-etal-2022-better,misra-mahowald-2024-language,Potts_2024}.
However, existing approaches remain limited for fine-grained linguistic analysis of constructions.
They typically do not enable comprehensive and systematically controlled comparisons across models and constructions.
Most focus on a single construction~\cite{weissweiler-etal-2022-better,misra-mahowald-2024-language,Potts_2024}, making it difficult to analyze relations among constructions or connect constructional understanding to other linguistic abilities.
In addition, evaluation formats such as natural language inference (NLI)~\cite{mackintosh-etal-2025-evaluating,scivetti-etal-2025-beyond} typically present a single sentence pair with a label and require a direct prediction.
In such settings, the linguistic cues driving the model’s judgments are often unclear.
Moreover, these tasks are not designed to support systematic comparison across models that span from developmentally plausible models to modern large language models (LLMs).
Smaller models often struggle with prompt interpretation or label classification.

In this work, we introduce the Linguistic Minimal-Pair Benchmark for Evaluating Constructional Understanding in Language Models (CxMP), which evaluates whether language models interpret the semantic relationships and participant roles implied by constructions.
For example, consider the let-alone construction shown in Figure~\ref{fig:overview} (``He can't help Lisa, let alone Emma.''), humans can infer that helping Emma is more difficult than helping Lisa.
We go beyond existing evaluations focused on the well-formedness of sentences and instead examine whether language models grasp the meanings encoded in linguistic form.
CxMP covers nine constructions, uses a controlled minimal-pair design to isolate linguistic cues, and supports evaluation across models ranging from developmentally plausible small models to large-scale LLMs.

We evaluate a range of language models that differ in model size, data scale, and training objectives.
We find that developmentally plausible small models make few correct judgments, while even the recent 70B-scale open models still struggle with some constructions, indicating substantial room for improvement in construction-level understanding.
Factors influencing performance vary across construction types: formally complex constructions are more affected by model size, whereas argument-structure constructions are more sensitive to instruction tuning.

To demonstrate the usefulness of our dataset for comparison with human language acquisition, we further analyze learning trajectories and pseudoword generalization.
Unlike BLiMP, which evaluates sentence acceptability, CxMP tests whether models interpret meanings encoded in syntactic form. 
Models reach high BLiMP accuracy early in training, whereas CxMP performance improves gradually.
A fine-grained analysis of CxMP reveals stage-like learning transitions from near-random responses to heuristic use and eventual stabilization.
Under a more controlled condition where lexical items are replaced with pseudowords, models, unlike humans, show a substantial decline in constructional meaning interpretation.

Taken together, these findings highlight the need for evaluation beyond traditional grammatical acceptability to capture the semantic contributions of constructions.
They further demonstrate that the proposed dataset provides a foundation for future research on constructional understanding.

\section{Motivations}
In this section, we outline four key properties that an evaluation dataset should satisfy to assess constructional understanding in language models effectively.
\paragraph{Model Agnostic}
Recent work has examined developmentally plausible small models to investigate language acquisition under limited data conditions~\cite{hu-etal-2024-findings}.
Evidence from large language models, however, shows that generalization can arise even without strong constraints, driven by inherent simplicity biases that offer insights for linguistic theory~\cite{Futrell_2025}.
Most existing studies rely on NLI~\cite{mackintosh-etal-2025-evaluating,scivetti-etal-2025-beyond}, where success presupposes understanding task-specific notions such as entailment and contradiction, making zero-shot evaluation difficult for smaller models that struggle with prompt interpretation and label classification.
Fine-tuning further obscures the boundary between pretrained and newly learned knowledge.
Hence, there is a need for benchmarks that enable consistent evaluation across both small and large language models.

\paragraph{Comprehensive across Constructions}
Most prior studies focus on individual constructions such as the Comparative Correlative~\cite{weissweiler-etal-2022-better}, AANN~\cite{misra-mahowald-2024-language}, or Preposing in PiPP~\cite{Potts_2024}, examining how language models recognize, reason about, and generalize over these specific patterns.
However, Construction Grammar encompasses a wide range of constructions, from lexically grounded substantive ones to highly abstract schematic ones.
A benchmark that systematically covers diverse constructions enables more comprehensive analyses of the relationships among constructions and their connections to other aspects of linguistic knowledge.

\paragraph{Controlled Design}
Datasets such as NLI and CoLA, which consist of sentences paired with labels, allow for diverse forms of evaluation but risk inducing shallow heuristics or surface-level biases in model judgments.
Moreover, the rationale for labeling is often implicit, making it unclear which linguistic cues the model relies on to make its judgments.
For example, sentences with high vocabulary overlap are more likely to be classified as entailment, whereas those containing negation tend to receive negative labels.
To mitigate such issues, a minimal-pair design that isolates controlled differences is desirable, enabling precise evaluation of which linguistic factors models are actually sensitive to.
Furthermore, common nouns, while natural and frequent, often exhibit strong lexical associations; for instance, teacher co-occurs with teach far more often than with learn.
To control for such lexical biases, it is important to introduce multiple noun variants and systematically alternate the roles or relationships of participants (e.g., swapping subject and object), allowing a more fine-grained analysis of models' underlying biases.

\paragraph{Multiple Sentence}
Single-sentence minimal pairs are effective for evaluating the formal naturalness of sentences but are insufficient for assessing the semantic contributions of constructions.
A multi-sentence design is therefore necessary.
Among existing datasets adopting such a design, EWoK serves as a representative example: it evaluates world knowledge through pairs consisting of a context describing a situation and a target proposition whose plausibility is judged relative to that context.
However, benchmarks that apply this framework to the study of language acquisition in language models from a linguistic perspective remain limited.
Since understanding a sentence's meaning may involve inferences drawn from its surrounding context during training, adopting a multi-sentence minimal-pair design enables investigation of how well language models capture semantic relations between sentences

In this work, we construct CxMP, a dataset that integrates all of these motivations into a unified framework.

\section{CxMP}

\begin{table*}[t]
    \centering
    \small
    \renewcommand{\arraystretch}{0.75}
    \begin{tabular}{cll} 
    \toprule
    Construction & Constructional Example & Diagnostic Sentence \\ 
    \midrule
    \multirow{2}{*}{\textsc{Let-alone}} & \multirow{2}{*}{He can't help N1, let alone N2.} &  Helping N2/*N1 is more difficult than helping N1/*N2. \\
      &  & Helping N1/*N2 is easier than helping N2/*N1. \\
    \cmidrule(lr){1-3} 
    \multirow{2}{*}{\textsc{Causative-with}}  & \multirow{2}{*}{N1 covered N2 with a coat.} & The one who provided the coat is N1/*N2. \\
      & & The one who received the coat is N2/*N1. \\
     \cmidrule(lr){1-1} \cmidrule(lr){2-2} \cmidrule(lr){3-3} 
    \multirow{2}{*}{\textsc{Way-manner}} & \multirow{2}{*}{N1 pushed his way into the room.} & He did so while moving/*staying. \\
      &  & He did so with/*without effort.\\
     \cmidrule(lr){1-1} \cmidrule(lr){2-2} \cmidrule(lr){3-3} 
    \multirow{7}{*}{\shortstack{\textsc{Comparative-}\\\textsc{correlative (\textsc{CC})}}}  
      & The harder you work, & \multirow{3}{*}{The one who becomes stronger is N1/*N2.} \\
      & the stronger you become. &  \\
      & N1 works harder than N2. & \\
    \cmidrule(lr){2-2} \cmidrule(lr){3-3}
      & The stronger you become, & \multirow{3}{*}{The one who works harder is N1/*N2.} \\
      & the harder you work. &  \\ 
      & N1 becomes stronger than N2. & \\
    \cmidrule(lr){1-1} \cmidrule(lr){2-2} \cmidrule(lr){3-3} 
     \multirow{2}{*}{\textsc{Conative}}  & \multirow{2}{*}{N1 kicked at the ball.} & Whether she succeeded in kicking the ball is unclear/*clear. \\
      & &  Whether the ball was affected is unclear/*clear. \\
     \cmidrule(lr){1-3}
    \multirow{2}{*}{\textsc{Ditransitive}}  & \multirow{2}{*}{N1 bought N2 a gift.} & The one who gave the gift is N1/*N2. \\
      &  & The one who received the gift is N2/*N1. \\
    \cmidrule(lr){1-1} \cmidrule(lr){2-2} \cmidrule(lr){3-3} 
    \multirow{2}{*}{\textsc{\textsc{Caused-motion}}}  & \multirow{2}{*}{N1 pulled N2 out of the water.} & The one who moved her is N1/*N2. \\
      &  & The one who got moved is N2/*N1. \\
    \cmidrule(lr){1-1} \cmidrule(lr){2-2} \cmidrule(lr){3-3}  
     \multirow{2}{*}{\textsc{Resultative}}  & \multirow{2}{*}{N1 pushed N2 awake.} & The one who caused her to become awake is N1/*N2. \\
      &  & The one who became awake is N2/*N1. \\
    \cmidrule(lr){1-1} \cmidrule(lr){2-2} \cmidrule(lr){3-3}
     \multirow{2}{*}{\shortstack{\textsc{Intransitive-}\\\textsc{motion}}}  & \multirow{2}{*}{N1 ran into N2.} & The one who moved is N1/*N2. \\
      &  & The one who he moved toward is N2/*N1. \\
    \bottomrule
    \end{tabular}
    \caption{Constructions that we focus on. * indicates implausible versions.}
    \label{tab:constructions}
\end{table*}

We assume that when a model correctly identifies the meaning encoded in a construction and the roles or relations of its entities, it demonstrates constructional understanding through successful form–meaning mapping.
CxMP consists of sentence pairs (Evaluation Pairs), each comprising a sentence containing a target construction (Constructional Example) and a sentence probing the meaning mapped to that construction (Diagnostic Sentence).
Examples for each construction are shown in Table~\ref{tab:constructions}, where N1 and N2 denote the entities involved.
Each Diagnostic Sentence has two minimally different versions, plausible and implausible.
Each construction also has two variants (A/B) that probe the construction's implied meaning and relational roles from different perspectives.
To maintain temporal consistency, all sentences are written in the past tense.

\subsection{Constructions}
In this work, following \citet{bonial-tayyar-madabushi-2024-construction}, we adopt nine types of constructions.
Examples of these constructions are shown in Table~\ref{tab:constructions}.
The constructions range from partially fixed patterns to fully schematic ones.
\label{appendix:constructions}
\paragraph{\textsc{Let-alone}}
In the fixed let alone construction, the two conjuncts represent different points on the same scalar dimension, where the second clause denotes a stronger degree・typically a stronger form of negation, than the first.
This construction has been analyzed in detail by Fillmore~\cite{fillmore1988regularity} and others, as it exhibits unique properties combining a negative polarity environment and a comparative relation.
In this dataset, we assume that actions at a higher point on the negative scale are harder to perform and design sentences that ask which action is more difficult.
While a wide range of syntactic elements (nouns, verbs, clauses) can appear around let alone, we focus on the simplest and most controllable case: nominal comparison.
Because the construction requires a licenser for negative polarity, we explicitly include negation in the stimuli using negative auxiliaries such as ``cannot'' and ``don't''.
Although \citet{bonial-tayyar-madabushi-2024-construction} includes the \textsc{Much-less}, it differs from let alone only in surface form and is therefore omitted here.

\paragraph{\textsc{Causative-with}}
This construction is characterized by a \textit{with}-phrase that follows the direct object.
The direct object denotes the affected entity, and the \textit{with}-phrase indicates what is applied or provided to it.
In our dataset, we restrict both the subject and the object to human nouns, allowing us to test whether the model can correctly interpret the supply–recipient relation between entities.

\paragraph{\textsc{Way-manner}}
This construction expresses movement achieved by creating or carving out one's own path, often despite external difficulty or through indirect means~\cite{goldberg1995constructions}.
Formally, a fixed phrase one's way appears in the direct-object position, followed by a prepositional phrase that indicates the path.
To evaluate the understanding of this construction, we design sentences asking whether the event involves motion or rest and whether it is carried out with strong intentionality.

\paragraph{\textsc{Comparative-correlative}}
This construction has been extensively studied as one of the most challenging phenomena to explain within the generative framework and as one of the most prominent constructions in English~\cite{fillmore1986varieties,HOFFMANN_2017}.
Recent probing studies have also examined how language models understand this construction~\cite{weissweiler-etal-2022-better}.
It consists of two clauses, both beginning with the + comparative forms of adjectives or adverbs~\cite{GOLDBERG2003219}.
The second clause functions as a dependent variable determined by the first, often implying causal or temporal relationships between the two.
Following \citet{weissweiler-etal-2022-better}, our dataset uses Constructional Examples in which the first clause establishes a relational condition between entities, and the subsequent sentence asks which entity exhibits the resulting dependent property.

\paragraph{\textsc{Conative}}
The following five constructions are those identified by \citet{goldberg1995constructions} as representative argument-structure constructions.
The conative construction features a direct object preceded by at, indicating that the agent attempts or makes an effort to act upon the target, without implying success.
In our dataset, half of the Constructional Examples instantiate the conative construction and the other half use standard transitive sentences.
In the Diagnostic Sentences, we evaluate on whether models can distinguish that in conative cases the result of the action is uncertain, whereas in transitive cases the action is successful and the object is affected.

\paragraph{\textsc{Ditransitive}}
This construction follows the schema SUBJ V OBJ1 OBJ2, expressing that the subject transfers or gives OBJ2 (the theme) to OBJ1 (the recipient).
We evaluate whether models capture this transfer-of-possession meaning by asking who gave and who received the item in the sentence.

\paragraph{\textsc{Caused-motion}}
This construction expresses that the subject (causer) causes the theme to move along a path indicated by an oblique phrase.
Formally, it follows the schema SUBJ V OBJ OBL, where the oblique phrase always denotes direction or path.
We evaluate whether models correctly assign causal semantic roles by asking Diagnostic Sentences, such as who caused the motion and who was moved.
\paragraph{\textsc{Resultative}}
This construction follows the pattern SUBJ V OBJ XCOMP, indicating that the subject's (agent's) action causes the object (patient) to undergo a change of state (XCOMP).
The XCOMP is typically an adjective phrase (e.g., flat, open) or prepositional phrase denoting the resulting state.
We evaluate whether models can identify which participant underwent the change and which one caused it, thereby testing their understanding of the causal and resultative semantics inherent to the construction.

\paragraph{\textsc{Intransitive-motion}}
This construction follows the schema SUBJ V OBL and denotes that the subject (agent) moves autonomously.
In our dataset, the oblique phrase also denotes an animate entity, and Diagnostic Sentences ask which participant moved and which one the mover arrived at, assessing whether the model understands directionality and goal relations.

\subsection{Data Generation with LLMs}
We automatically generate Constructional Examples for each construction using gpt-5-chat-latest based on predefined templates designed to elicit a corresponding Diagnostic Sentence.\footnote{Detailed generation prompts, templates, and filtering criteria are provided in Appendix~\ref{appendix:dataset_generation}.}
For each construction, we prompt the model to fill slot variables (e.g., subject, oblique) with contextually appropriate words, and use multiple random seeds to enhance lexical diversity.
We then filter the generated sentences and pair them with corresponding plausible and implausible diagnostic statements, ensuring that the final dataset includes only semantically coherent form–meaning mappings.
We manually inspect the resulting dataset to ensure its quality and validity.
In total, we finally create 43k Evaluation Pairs.

\paragraph{Human Data Validation}
We conduct a human validation study using Prolific with native English speakers.
Participants read a preceding sentence and select the continuation that is more plausible.
We collect three independent annotations per item.
Across 896 items, the majority-vote accuracy reaches 96.65\%, and full agreement among annotators reaches 83.59\%, suggesting that the automatically generated items are generally reliable.
The number of items varies by construction (typically 112, and 56 for constructions without switched variants). Detailed per-construction results and annotation statistics are provided in Appendix~\ref{appendix:data-validation}.

\paragraph{Entities and Generation Procedure}
We used four types of names that represent entities, \textit{Female name}, \textit{Male name}, and \textit{Name+alphabet} (e.g., \textit{Name B}), and \textit{common noun}.\footnote{Details of name selection and switching procedures are provided in Appendix~\ref{appendix:dataset_generation}.}
While common nouns represent the most natural expressions, proper-name variants mitigate lexical biases, and name–alphabet forms provide the most controlled condition.
For constructions involving two entities, we also created switched versions by reversing their positions to test whether models rely on shallow heuristics.
They correspond to the sentence swapping N1 and N2 in Table~\ref{tab:constructions}.

\section{Experiment}
\subsection{Settings}

\paragraph{Models}

\begin{figure*}[tbp]
\centering
\includegraphics[width=\linewidth]{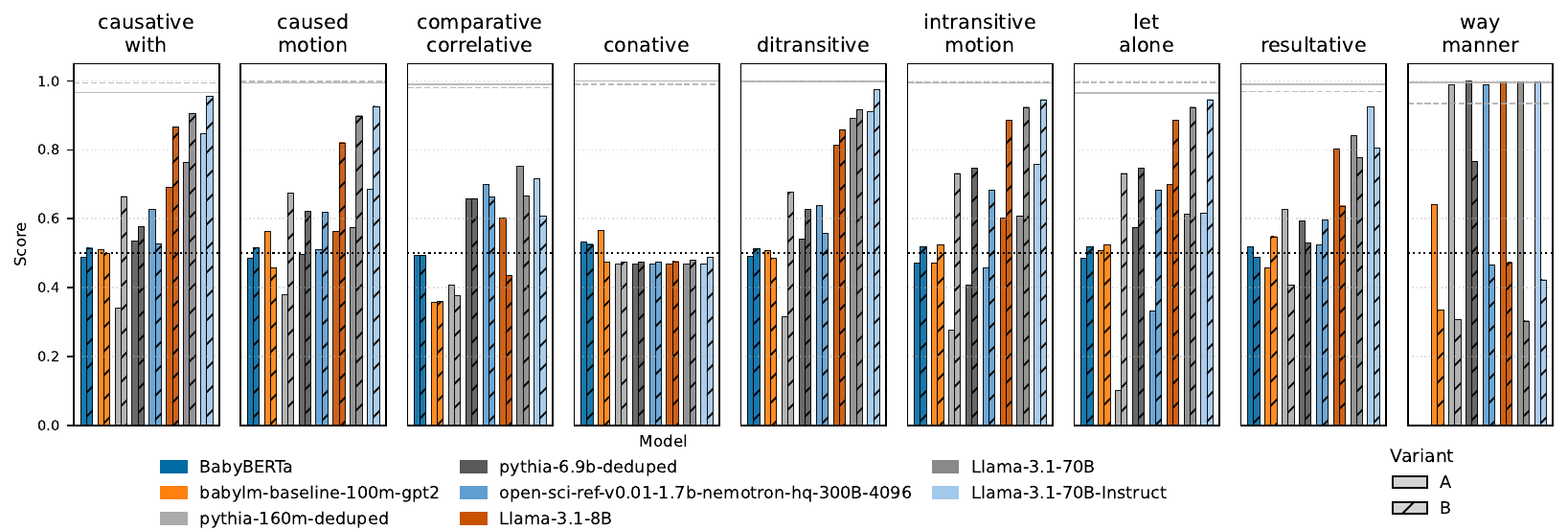}
\caption{Scores of each language model across constructions and variants. Solid and dashed lines at the top of the figure represent the scores of GPT-5 for variants A and B, respectively.}
\label{fig:results}
\end{figure*}

We use developmentally appropriate models from the BabyLM Challenge (10M/100M tokens)~\cite{charpentier2025babylmturns3papers}, trained on child-directed corpora such as CHILDES, and three BabyBERTa models trained with different random seeds~\cite{huebner-etal-2021-babyberta}.
To examine data size effects, we use Open-sci-ref-0.01~\cite{nezhurina2025opensciref001openreproduciblereference}, which provides 1.3B and 1.7B models trained on C4 (50B/300B tokens) and Nemotron (300B/1T tokens), respectively.
For model size effects, we use the Pythia suite~\cite{biderman2023pythia}, which includes ten models from 14M to 12B parameters.
To compare training objectives, we contrast causal and masked language modeling using GPT-2-medium, RoBERTa-base/large, and Pythia-410m.
Their sizes are comparable, while data sizes increase in the order GPT-2 (40G) < RoBERTa (160G) < Pythia (825G).
For larger open LLMs, we use Llama-3.1~\cite{grattafiori2024llama3herdmodels} (8B/70B, base and instruction-tuned) to assess the effect of instruction tuning.
Finally, we include GPT-5\footnote{\url{https://openai.com/index/introducing-gpt-5/}}
, a widely used closed-source commercial model, as an approximate upper bound for open models.

\paragraph{Evaluation}
To evaluate whether each language model grasps the construction and its meaning, we computed the probabilities of the Plausible Evaluation Pair and Implausible Evaluation Pair for each set.
We use length-normalized log probability for CLM and pseudo log-likelihood~\cite{salazar-etal-2020-masked} for MLM.
In contrast, GPT-5 is not directly accessible to probabilities; we estimate its likelihoods through prompting.
Details of the evaluation procedure are in Appendix~\ref{appendix:evaluation}.

\subsection{Results}
\label{subsec:results}
\paragraph{Overall}
We evaluate eight models of varying sizes from the series used in this study, as shown in Figure~\ref{fig:results}.
Small-scale models trained on developmentally appropriate corpora perform at a near-chance level.
While previous work has shown that even small models can perform reasonably well on formal grammatical benchmarks such as BLiMP, our results indicate that understanding constructional meaning is considerably more challenging.
In contrast, the large closed-source GPT-5 accurately interprets most constructions.
However, even open models with 70B parameters continue to struggle with certain constructions, and some remain entirely unsolved.
Mid-sized models, in particular, show a mix of very high and very low accuracies across constructions, suggesting reliance on specific heuristics or biases.
A detailed analysis of these biases is provided in Section~\ref{sec:bias}.

\paragraph{Data/Model Size}
\begin{figure}[tbp]
\centering
\includegraphics[width=\linewidth]{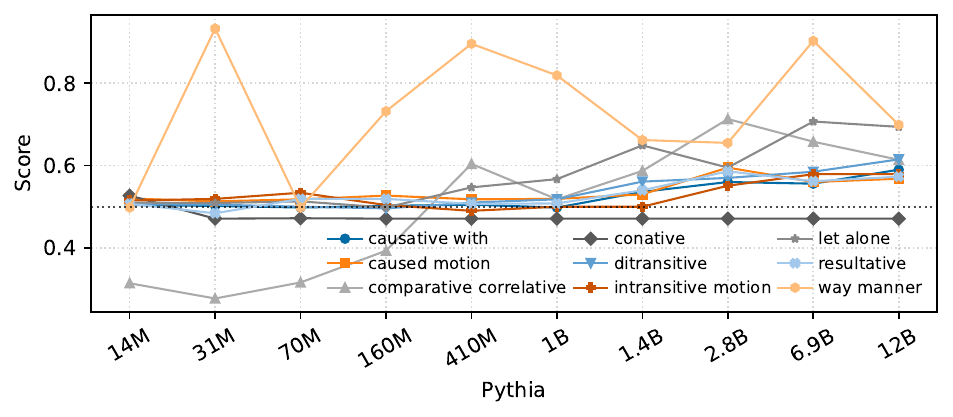}
\caption{Scores of Pythia models by model size.}
\label{fig:pythia_results}
\end{figure}

Figure~\ref{fig:pythia_results} shows the accuracy of each construction across ten model sizes in the Pythia series.
Overall, except for the \textsc{Conative}, most constructions exhibit a clear upward trend in scores as model size increases.
In particular, non-standard constructions such as \textsc{Let-alone} and \textsc{CC} show a strong dependence on model expressivity, with substantial performance improvements observed only in larger models.
For the remaining constructions, scores begin to rise around the 1.4B parameter scale, while smaller models remain near chance level.
In contrast, \textsc{Way-manner} shows large fluctuations across model sizes, with no consistent trend.

\begin{figure}[tbp]
\centering
\includegraphics[width=\linewidth]{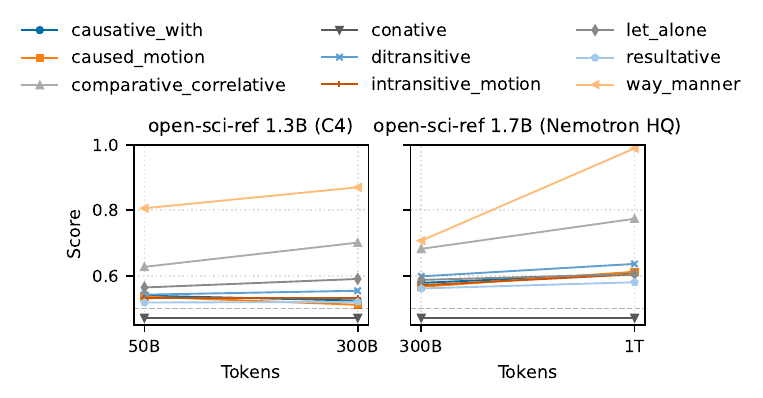}
\caption{Scores of open-sci-ref models by data size.}
\label{fig:opensci_results}
\end{figure}

Figure~\ref{fig:opensci_results} presents results from the Open-sci-ref-0.01 models trained on different data scales.
On average, performance improves with larger training datasets, though the gain from 50B to 300B tokens remains limited.
At the level of individual constructions, most show minimal change, or even slight decreases, between 50B and 300B.
However, all constructions exhibit score increases between 300B and 1T, suggesting that, under this setup, larger-scale training data substantially contribute to improvements in constructional meaning understanding.

\paragraph{Instruction Tuning}
\begin{figure}[tbp]
\centering
\includegraphics[width=\linewidth]{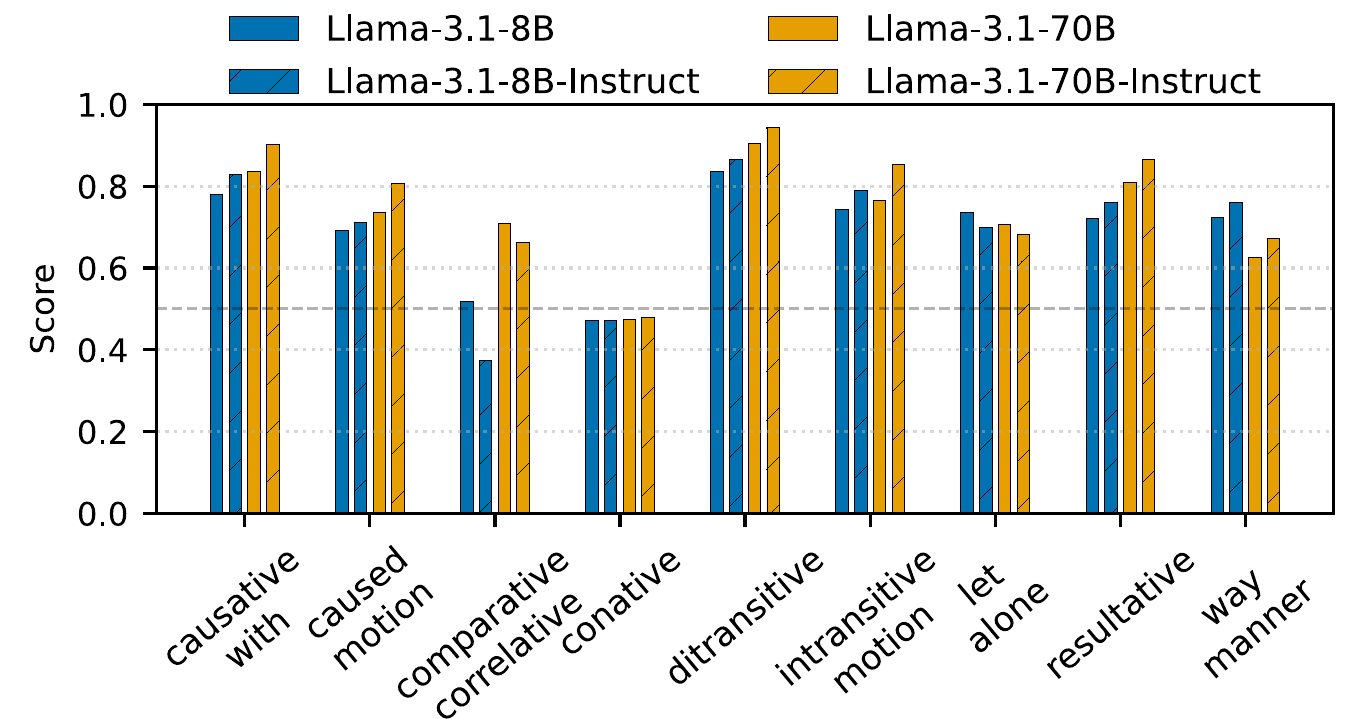}
\caption{Scores of Llama3.1 base/instruction models.}
\label{fig:llama31_results}
\end{figure}
Figure~\ref{fig:llama31_results} shows the scores of the base and instruction-tuned variants of Llama3.1 8B/70B.
Overall, the instruction-tuned models tend to achieve higher scores than their base counterparts.
However, the magnitude of this effect varies across constructions: while consistent improvements are observed for argument-structure constructions, non-standard constructions such as \textsc{Let-alone} and \textsc{CC} show a decrease in performance.
These results suggest that instruction tuning does not necessarily enhance all aspects of constructional meaning understanding.

\paragraph{Objectives}

Table~\ref{tab:clm_mlm_results} compares RoBERTa and CLM models of comparable size.
Although RoBERTa-large is slightly larger, it outperforms Pythia despite Pythia being trained on over five times more data.
Similarly, RoBERTa-base achieves higher scores than the Pythia model, exceeding it in both model and data size.
Overall, within the examined range, MLMs tend to outperform CLMs.
Recent theoretical work~\cite{pmlr-v235-zhang24m} suggests that MLMs induce richer co-occurrence patterns and stronger semantic associations, enhancing classification performance, whereas their fixed masking rate limits generalization in generative tasks with variable-length inputs.
Although our task differs from probability-based classification, the results indicate that MLMs may exploit bidirectional context to more effectively integrate form and meaning.
Consistent with this trend, several BabyLM Challenge systems~\cite{hu-etal-2024-findings} have adopted hybrid MLM–CLM objectives~\cite{yu-etal-2024-antlm,charpentier-samuel-2024-bert}.
A systematic exploration of which linguistic abilities particularly benefit from the MLM objective remains a promising direction for future work.

\begin{table}[tbp]
    \small
    \renewcommand{\arraystretch}{0.75}
    \centering
    \begin{tabular}{@{}lcccc@{}}
        \toprule
        Model & Model sz. & Data sz. & Obj. & Score \\
        \midrule
        RoBERTa-base  & 125M & 160G & MLM & 0.573 \\
        GPT-2-medium     & 345M & 40G & CLM & 0.537 \\
        RoBERTa-large   & 355M & 160G & MLM & \textbf{0.674} \\
        Pythia-410m     & 410M & 825G & CLM & 0.544 \\
        \bottomrule
    \end{tabular}
    \caption{Scores of CLM \& MLM models. Obj. indicates objective and Sz. indicates size.}
    \label{tab:clm_mlm_results}
\end{table}

\section{Analysis}

\subsection{Learning Trajectories}
\label{sec:trajectory}
Our work investigates the linguistic ability to interpret meanings implied by linguistic forms, an ability not captured by grammatical acceptability alone.
This section traces the development of grammatical acceptability and constructional meaning understanding across training and data size.
We hypothesize that grammatical acceptability improves earlier in training, as it often depends on relatively shallow statistical patterns.
For instance, in number agreement, a model that learns to associate the number feature of the subject with the verb can correctly prefer ``The cats annoy Tim.'' over ``*The cats annoys Tim.''
In contrast, many constructions in CxMP require grasping the meaning encoded by the form itself, which likely demands more training data and emerges later in learning.

\paragraph{Settings}
We use the base versions of OLMo2-7B/13B, which provide publicly available checkpoints~\cite{olmo20242olmo2furious}.
We select 11 representative checkpoints from Stage 1 of OLMo2-13B and several checkpoints from Stage 2.
For OLMo2-7B, we adopt checkpoints that correspond to approximately the same cumulative token counts as those used for the 13B model.
For evaluation Datasets, we use BLiMP~\cite{warstadt-etal-2020-blimp-benchmark} in addition to CxMP.
BLiMP is a benchmark designed to evaluate the grammatical sensitivity of language models.
It consists of pairs of (un)grammatical sentences in 12 linguistic phenomena, and we compute scores for each phenomenon.
The scoring method is identical to that used for CxMP.

\paragraph{Results}
\begin{figure}[tbp]
  \centering

  \begin{subfigure}[t]{\linewidth}
    \centering
    \includegraphics[width=\linewidth]{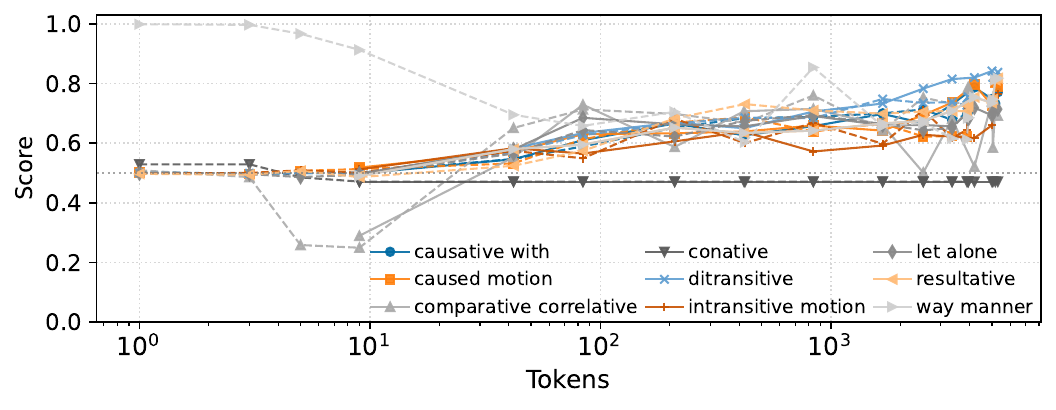}
    \caption{Scores on CxMP}
    \label{fig:results_cxmp}
  \end{subfigure}


  \begin{subfigure}[t]{\linewidth}
    \centering
    \includegraphics[width=\linewidth]{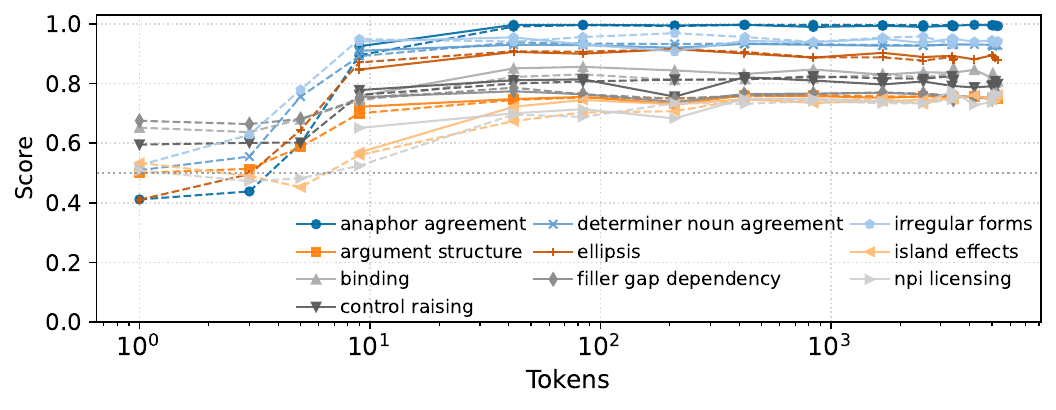}
    \caption{Scores on BLiMP}
    \label{fig:results_blimp}
  \end{subfigure}

  \caption{Scores of OLMo-2 on CxMP and BLiMP across training token counts (log scale). Solid/Dashed lines represent the 13B/7B model.}
  \label{fig:results_cxmp_blimp}
\end{figure}
Figure~\ref{fig:results_cxmp_blimp} shows the trajectories of BLiMP and CxMP scores (log scale) as a function of training token count.
For BLiMP, the models correctly solve roughly 80\% of linguistic phenomena by around 50B tokens, after which the scores plateau, showing only a minimal increase of about one to two points overall.
In contrast, CxMP scores continue to improve gradually from around 10B tokens until the final stage of training.
These results suggest that language models acquire formal linguistic knowledge, such as grammatical acceptability, relatively early in training, whereas mastering more complicated linguistic abilities, such as constructional meaning, requires longer-term learning.
Traditionally, model performance has been primarily assessed with BLiMP, and reaching a certain score has often been taken as evidence of sufficient linguistic competence.
However, the results from CxMP reveal that understanding not only the formal grammatical aspects of constructions but also their semantic and functional dimensions requires continued learning.

\subsection{Effect of Bias}
\label{sec:bias}
In Section~\ref{subsec:results}, we find that models trained on developmentally appropriate data sizes and domains remain near chance level on many constructions and variants.
In contrast, mid-sized models show large score gaps across variants, whereas larger models. This tendency is smaller in larger models.
This pattern suggests the presence of biases driven by shallow heuristics.
We hypothesize that such biases may appear not only between variants but also between switched and non-switched sentences.
This section analyzes how these biases evolve throughout training, aiming to identify when such heuristic-driven behaviors begin to emerge.

\paragraph{Definition of Bias}

(i) Bias between switched and non-switched sentences:
This bias arises when scores differ between switched and non-switched sentences within the same construction and variant.
For example, in ``Gianna left Nancy with a receipt. The one who provided the receipt is Gianna/*Nancy.'' the model correctly prefers the plausible evaluation pair.
However, in the switched version ``Nancy left Gianna with a receipt. The one who provided the receipt is Nancy/*Gianna.'' it still tends to choose ``Gianna.''
In such cases, the model may rely solely on the diagnostic sentence, mechanically selecting the subject most frequently associated with the verb.
This bias is likely more pronounced when the protagonists are common nouns, as switching reverses the typical agent–patient relation (e.g., The student taught the teacher English.).
(ii) Bias across variants:
This bias occurs when scores differ between variants of the same construction.
For instance, while the model correctly prefers the plausible evaluation pair in ``Gianna left Nancy with a receipt. The one who provided the receipt is Gianna/*Nancy.'' it still favors ``Gianna'' in ``Gianna left Nancy with a receipt. The one who received the receipt is Nancy/*Gianna.'' That is, it selects the implausible evaluation pair.
Such patterns suggest that the model may rely on shallow heuristics based on surface cues, such as choosing the nearest noun or consistently selecting the first noun in the sentence.

\paragraph{Quantification of Bias}
We evaluate each checkpoint of the same models used in Section~\ref{sec:trajectory} using the following metrics.
(i) Bias between switched and original sentences: For each construction and noun type, we computed the absolute difference between the mean scores of the original and switched variants (A vs. A\_switched, B vs. B\_switched).
(ii) Bias across variants: For each construction and noun type, we calculate the absolute difference between the mean scores of the two variants (A/A\_switched, B/B\_switched), where each variant's score was averaged over its original and switched forms.
We then average these values across constructions within each noun type.
Note that larger values would indicate stronger bias in the model.

\paragraph{Results}
\begin{figure}[tbp]
  \centering

  \begin{subfigure}[t]{\linewidth}
    \centering
    \includegraphics[width=\linewidth]{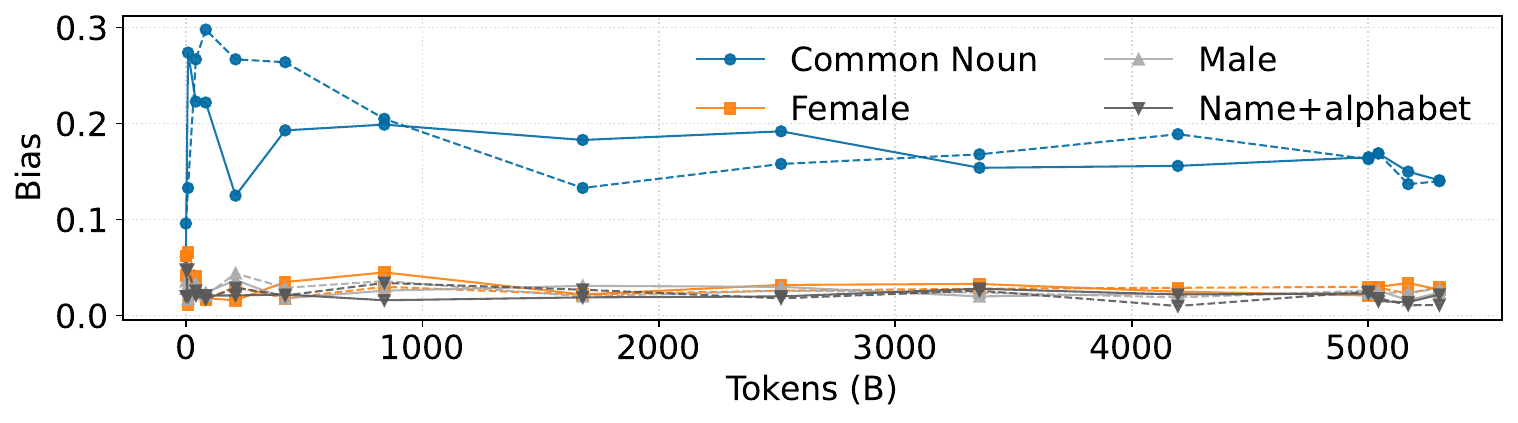}
    \caption{Bias b/w switched and non-switched sentences (;i)}
    \label{fig:diff_switch}
  \end{subfigure}

  \begin{subfigure}[t]{\linewidth}
    \centering
    \includegraphics[width=\linewidth]{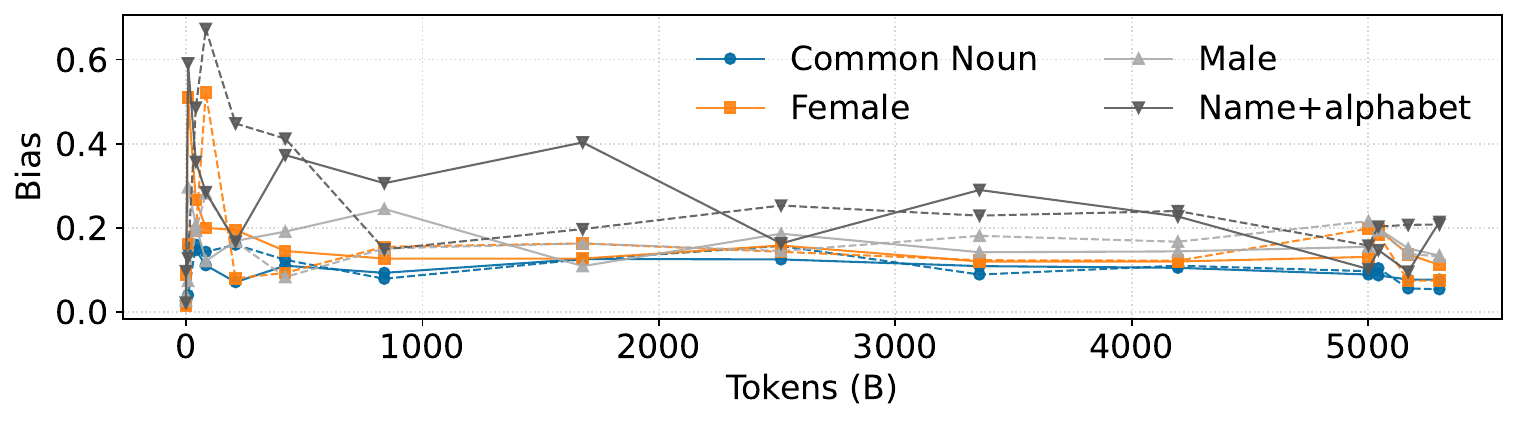}
    \caption{Bias across variants (;ii)}
    \label{fig:diff_variant}
  \end{subfigure}

  \caption{Biases of OLMo-2 on CxMP across training token counts (log scale).}
  \label{fig:diff_bias}
\end{figure}
Figures~\ref{fig:diff_switch} and~\ref{fig:diff_variant} show the score trajectories across checkpoints (training token count) for each noun type, corresponding to Type (i) and Type (ii), respectively.
In Type (i), this trend is most prominent for \textit{Common noun}s, while in Type (ii), it appears more broadly, especially for \textit{Name+alphabet}.
In both cases, the scores start low in the early training stages, rise thereafter, and gradually decline again toward the end.
Compared with the results in Section~\ref{sec:trajectory} (Figure~\ref{fig:results_cxmp}), this pattern suggests that the model initially responds at random without bias, then develops shallow heuristics, and finally converges toward more accurate judgments.
As expected, in Type (i), this tendency is particularly evident for Common nouns.
In contrast, in Type (ii), a similar pattern emerges most clearly for Name+alphabet.
This may indicate that the model becomes confused by low-frequency input formats and overly relies on nearby cues.
Overall, these results demonstrate that different kinds of biases can arise depending on the noun type and sentence structure being tested.
They also highlight the importance of incorporating diverse question formulations to comprehensively assess the model's linguistic knowledge.

\subsection{CxMP Using Pseudoword}
In theoretical linguistics, the meanings of constructions are theoretically defined and systematized~\cite{goldberg1995constructions}.
In contrast, \citet{Kako01082006} experimentally examines how speakers perceive semantic properties directly from constructions themselves.
He uses stimulus sentences such as ``The rom gorped the blickit to the dax.'', in which verbs and nouns are replaced with meaningless pseudowords, thereby controlling for lexical semantics and evaluating how humans interpret meaning solely from constructional form.
The results reveal significant correspondences between each construction and the semantic attributes theoretically assumed in prior work.
In this section, adopting a similar perspective, we examine how well language models can capture the correspondence between constructions and their meanings in sentences where the influence of lexical semantics is mitigated.
Specifically, using CxMP, we evaluate whether models can reconstruct meaning based on constructional form.

\paragraph{Settings}
We create the evaluation dataset used in the experiment by replacing the content words in CxMP with pseudowords.
Our analysis focuses on four major argument-structure constructions identified by Goldberg~\cite{goldberg1995constructions}: \textsc{Caused-motion}, \textsc{Intransitive-motion}, \textsc{Ditransitive}, and \textsc{Resultative}.
Among these, \textsc{Ditransitive} and \textsc{Intransitive-motion} overlap with the constructions examined by \citet{Kako01082006}.
We generate the pseudowords with Wuggy~\cite{keuleers2010wuggy}, a tool that produces phonotactically natural words (\textit{wug}s).
The detailed settings are in Appendix~\ref{appendix:wug_setting}.
We use Llama3.1–70B as the target model.

\paragraph{Results}
\begin{figure}[tbp]
\centering
\includegraphics[width=0.7\linewidth]{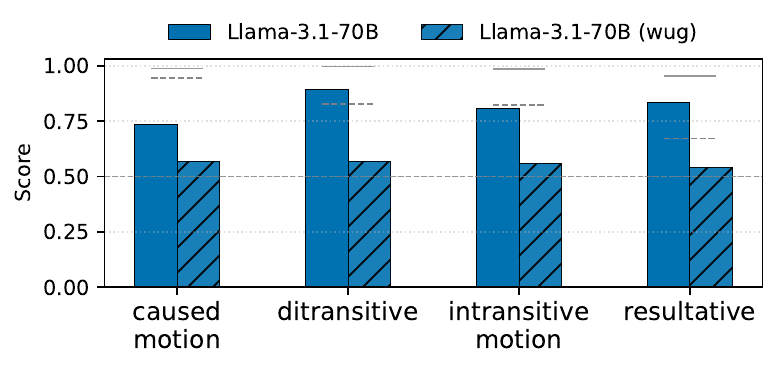}
\caption{Scores for pseudowords (\textit{wug}) and real words.}
\label{fig:pseudo_results}
\end{figure}

Figure~\ref{fig:pseudo_results} shows the scores for each construction using either pseudowords (\textit{wug}s) or the original words.
Across all constructions, scores drop markedly when original words are replaced with pseudowords, indicating that the model struggles to infer meaning solely from constructional cues without relying on lexical content.
Nevertheless, scores remain slightly above chance, suggesting that the model retains some sensitivity to constructional information.
By construction type, \citet{Kako01082006} reported that the presence or absence of closed-class words (e.g., to, that) had little effect on performance.
Among the constructions tested here, \textsc{Caused-motion} and \textsc{Intransitive-motion} include such closed-class words, and their pseudoword–original differences are smaller than those of other constructions, with overall scores remaining stable.
The pseudowords used are phonotactically natural and morphologically plausible, potentially sharing letters with the original words and thus offering weak cues.
Following prior work, we also retained determiners and inflectional suffixes such as -ed and -ing, which may provide part-of-speech cues that facilitate construction identification.
Future work should further examine which linguistic elements contribute to identifying constructions under varying levels of lexical control.

\section{Conclusion}
We present CxMP, a Construction Grammar–based benchmark for evaluating how language models grasp meanings encoded by constructions.
Results indicate that there remains considerable room for improvement in constructional meaning understanding, not only in developmentally plausible small models but also in modern LLMs.

\section*{Limitations}
While CxMP provides a controlled and interpretable framework for evaluating constructional understanding in language models, several limitations remain.
First, although the dataset covers a broad range of constructions, it focuses primarily on English and on a limited subset of construction types. Extending the benchmark to other languages or to more diverse construction families would allow testing whether the observed developmental patterns generalize across typologically different systems.
Second, the diagnostic design relies on written, isolated sentences. Real-world language understanding involves richer discourse and pragmatic contexts, which may influence how constructions are interpreted. Incorporating contextualized or multi-turn settings would improve ecological validity.
Third, while pseudoword experiments mitigate lexical bias, they may also introduce unnatural stimuli that underestimate model competence. Future work could explore alternative methods for lexical control that preserve naturalness.
Finally, although CxMP evaluates a wide range of models, it remains static. Coupling this benchmark with dynamic evaluation protocols, such as probing internal representations or testing fine-tuned adaptations, could reveal deeper insights into how constructional meaning emerges and is represented within models.

\section*{Acknowledgments}
We would like to express our gratitude to the anonymous reviewers who provided many insightful comments that have improved our paper. 
This work was supported by JSPS	KAKENHI Grant Numbers JP25KJ1824 and JP25K21281, JST BOOST Grant Number JPMJBY24D9, and JST FOREST Grant Number JPMJFR232R.

\bibliography{custom}

\appendix

\section{License}
\begin{enumerate}
    \item wuggy: MIT license
    \item BLiMP: CC-BY
    \item SSA Baby Names: CC0/Public Domain
    \item Pythia: Apache-2.0
    \item Llama-3.1: Meta Llama 3.1 Community License
    \item OLMo-2.0: Apache-2.0
    \item BabyLM baselines: Apache-2.0
    \item RoBERTa: MIT license
    \item BabyBERTa: MIT license
    \item GPT2: MIT license
    
\end{enumerate}

\section{Intended-Use and Access-Conditions Compliance}

We use each external artifact within the scope of its intended use and access conditions:
Research benchmarking and analysis only (no end-user deployment).
Derivative datasets generated from public sources (SSA names) and LLM outputs are released solely for research/education.
For license-restricted models (e.g., Llama-3.1), we comply with the Community License terms and do not redistribute weights.


\section{Dataset Generation}
\label{appendix:dataset_generation}
\label{appendix:dataset_design}
First, we generate Constructional Examples for each construction using a GPT5
The model used for generation is gpt-5-chat-latest.
For each construction, we instruct the model to produce sentences based on predefined templates so that a corresponding Diagnostic Sentence naturally follows.
For example, in the \textsc{Causative-with}, we use the following template:``\{Subj; N1; human common noun (gender-neutral)\} \{V\} \{Obj; N2; human common noun (gender-neutral)\} \{Obl; with-PP\}''
We then attach reference Diagnostic Sentences such as ``The one who provided \{Obl's noun\} is N1.'' and ``The one who received \{Obl's noun\} is N2.'', and instruct the LLM to fill each curly brace with contextually appropriate words.
For each construction, we generate 50 sentences using at least eight different random seeds and remove duplicates.
In constructing the prompts for sentence generation, we paid careful attention to several design considerations. Specifically, we instructed the model to (i) use clear and simple language that is commonly used, (ii) avoid semantically unnatural or implausible sentences, (iii) include verbs whose interpretation depends on the construction rather than only on the verb itself, while ensuring that the reference sentences connect naturally, (iv) avoid reusing words that have already appeared in the output while preserving the intended meaning of the construction, (v) refrain from including adverbs or modifiers not explicitly specified in the template, and (vi) generate only affirmative sentences (except for \textsc{Let-alone}).
We then classify the generated sentences into three categories, Prototypical, Extended (Peripheral), and Unacceptable (Anomalous), and exclude those labeled as Unacceptable.
Next, we generate Diagnostic Sentences.
For each construction and variant, we prepare templates for Plausible Diagnostic Sentences and Implausible Diagnostic Sentences (e.g., ``The one who provided \{Obl's noun\} is \{n1\}.'') and replace the corresponding words in the Constructional Examples with curly braces.
Finally, we ask the LLM to judge whether each sentence and its diagnostic counterpart are semantically aligned, and we compute the final scores only for pairs confirmed to be semantically consistent.
To reduce the effect of shallow heuristics~\cite{mccoy-etal-2019-right}, Diagnostic Sentences are designed so that each of their words occurs an equal number of times across all Constructional Examples, including those without the corresponding word.
We finally create approximately 300 Evaluation Pairs × 9 constructions × 2 variants × 4 noun types × (with and without switching)

\section{Huma Data Validation}
\label{appendix:data-validation}
We conduct a human validation study to assess the plausibility of automatically generated items. 
We recruit 168 participants via Prolific and restrict recruitment to native English speakers with an approval rate of at least 97\%. 
Participants read a preceding sentence and select the continuation that is more plausible, and we collect responses using Google Forms. We collect three independent annotations per item, and each participant completes 16 items (approximately 10 minutes). 
Each construction contains112items, except \textsc{way\_manner} and \textsc{conative}, which contain 56 items because switched variants are not applicable. 
We include three types of attention checks: (1) a simple subject–verb agreement question, (2) a shuffled sentence detection task, and (3) an instruction-following question matching the example format, and we retain only participants who pass all checks.
Participants receive 1.5 GBP (excluding platform fees). We report majority-vote accuracy (the proportion of items where at least two of three annotators select the correct continuation) and full agreement rate (the proportion of items where all three annotators give the same answer).

\paragraph{Results}
Overall, the human validation results indicate that the automatically generated items are generally plausible. Across 896 items, the majority-vote accuracy reaches 96.65\%, and full agreement among annotators reaches 83.59\%. Most constructions achieve majority accuracy above 97\%. Some constructions (e.g., \textsc{conative}) show lower agreement, indicating that judgments for these items may be more variable. These statistics provide an additional sanity check for the generated dataset.

\begin{table}[t]
\centering
\small
\tabcolsep=5pt
\begin{tabular}{@{}lrrr@{}}
\toprule
Construction & \#Items & Maj. (\%) & Agr. (\%) \\
\midrule
\textsc{Let\_alone} & 112 & 97.32 & 66.07 \\
\textsc{Causative\_with} & 112 & 97.32 & 91.96 \\
\textsc{Way\_manner} & 56 & 98.21 & 80.36 \\
\textsc{CC} & 112 & 100.00 & 91.96 \\
\textsc{Conative} & 56 & 67.86 & 44.64 \\
\textsc{Ditransitive} & 112 & 100.00 & 89.29 \\
\textsc{Caused\_motion} & 112 & 97.32 & 89.29 \\
\textsc{Resultative} & 112 & 99.11 & 91.96 \\
\textsc{Intransitive\_motion} & 112 & 99.11 & 85.71 \\
\midrule
Overall & 896 & 96.65 & 83.59 \\
\bottomrule
\end{tabular}
\caption{Human validation results. Majority (Maj.) denotes the proportion of items where at least two of three annotators selected the correct continuation. Agreement (Agr.) denotes the proportion of items where all three annotators gave the same answer.}
\end{table}
 
\section{Evaluation Methods}
\label{appendix:evaluation}
To evaluate whether each language model can correctly judge the construction and its associated meaning, we computed the probabilities of the Plausible Evaluation Pair and Implausible Evaluation Pair for each set.
To reduce bias due to sentence length (i.e., token count), we use the length-normalized log probability, obtained by dividing the total log probability of the sentence by its number of tokens.
A model is considered correct when it assigns a higher probability to the Plausible Evaluation Pair, and the accuracy of these judgments was used as the score.
For MLMs, we compute probabilities using the pseudo log-likelihood method~\cite{salazar-etal-2020-masked}.
In contrast, for chat models such as GPT-5, where probabilities are not directly accessible, we estimate relative likelihoods through prompting.
We carefully designed the prompts so as not to give the model an advantage over direct probability computation; specifically, the model was instructed to choose the sentence that is higher likelihood.
To prevent reliance on local cues rather than holistic sentence comparison, the prompt explicitly required the model to evaluate the entire pair of sentences.
Each evaluation pair was presented individually, with the two options labeled as A and B.
To mitigate potential label bias (e.g., a tendency to consistently prefer one label such as A), the assignment of correct and incorrect sentences to the indices was randomized.

\section{Entities}
We define four types of entities: \textit{Common noun}, \textit{Female name}, \textit{Male name}, and \textit{Name+alphabet}.
The \textit{common noun} type consists of human-denoting nouns and represents the most natural variant.
However, certain nouns frequently co-occur with specific semantic roles, making it difficult to determine whether a model's correct answer reflects genuine understanding of the constructional meaning or mere reliance on lexical co-occurrence statistics.
For instance, when the verb is teach and the characters are teacher and student, the model may simply infer that the teacher is the agent and the student is the patient, as this pattern is highly frequent in the data.
The \textit{Female name} and \textit{Male name} variants maintain naturalness while reducing lexical co-occurrence bias.
Nonetheless, name-specific frequency effects may remain, for example, if Maria frequently appears as a giver in the training corpus.
The names are randomly selected from the top-10 entries (every decade) in the SSA Baby Names dataset\footnote{\url{https://www.ssa.gov/oact/babynames/}} from 1950 to 2020, after removing duplicates.
The \textit{Name+alphabet} type is the most artificial and controlled format.
We randomly sample an alphabetic label (e.g., A, B) and insert it into a template such as ``Name A.''
Because of its low frequency, this variant can reduce performance, particularly in smaller models with limited ability to generalize to unseen forms.

For cases involving two entities, we also create a switched version by reversing their positions.
This design helps test whether models rely on shallow heuristics, such as always choosing the first noun or the one closest to a particular syntactic position.
During sentence generation, we first create a baseline sentence using common nouns, then replace the entities with other types or switch their order as needed.

\section{Models}
OLMo2-13B is pretrained on 5T tokens and OLMo2-7B on 4T tokens.
The training of OLMo2 consists of two pre-training stages.
Stage 1 accounts for more than 90\% of the total training, while Stage 2 performs additional fine-tuning on high-quality data.
Specifically, the 7B model undergoes three runs of 50B tokens each, and the 13B model is further trained on 300B tokens.
The final released model is obtained by model souping, and therefore does not correspond to the final checkpoint.

\section{Generate Wug}
\label{appendix:wug_setting}
We input each original content word to Wuggy to obtain a corresponding pseudoword.
Following \citet{Kako01082006}, determiners and inflectional suffixes (e.g., -ed, -ing) are retained.
When Wuggy returns None as its output, we create a pseudoword by randomly generating alphabetic strings with the same number of entities as the original content word.

\end{document}